\documentclass[letterpaper]{article} % DO NOT CHANGE THIS
\usepackage{aaai2027}  % DO NOT CHANGE THIS
\usepackage{times}  % DO NOT CHANGE THIS
\usepackage{helvet}  % DO NOT CHANGE THIS
\usepackage{courier}  % DO NOT CHANGE THIS
\usepackage[hyphens]{url}  % DO NOT CHANGE THIS
\usepackage{graphicx} % DO NOT CHANGE THIS
\usepackage{svg}
\usepackage{natbib}  % DO NOT CHANGE THIS AND DO NOT ADD ANY OPTIONS TO IT
\usepackage{caption} % DO NOT CHANGE THIS AND DO NOT ADD ANY OPTIONS TO IT
\usepackage{tabularx}
\usepackage{algorithm}
\usepackage{algorithmic}

\usepackage{amsmath} 
\usepackage{subcaption}
\usepackage{amssymb}

\newcommand{\sysName}{\textit{AcoustiPrint}}
\usepackage{newfloat}
\usepackage{listings}
\DeclareCaptionStyle{ruled}{labelfont=normalfont,labelsep=colon,strut=off} % DO NOT CHANGE THIS
\floatstyle{ruled}
\newfloat{listing}{tb}{lst}{}
\floatname{listing}{Listing}
\title{Physics-Guided Neural Design of Manufacturable Acoustic Structures}
\author{
    Written by AAAI Press Staff\textsuperscript{\rm 1}\thanks{With help from the AAAI Publications Committee.}\\
    AAAI Style Contributions by Pater Patel Schneider,
    Sunil Issar,\\
    J. Scott Penberthy,
    George Ferguson,
    Hans Guesgen,
    Francisco Cruz\equalcontrib,
    Marc Pujol-Gonzalez\equalcontrib
}
\affiliations{
    \textsuperscript{\rm 1}Association for the Advancement of Artificial Intelligence\\
    1101 Pennsylvania Ave, NW Suite 300\\
    Washington, DC 20004 USA\\
    proceedings-questions@aaai.org
}

\title{AcoustiPrint: Physics-Guided Neural Design of Manufacturable Acoustic Structures}
\author {
    Shanmuka Sadhu\textsuperscript{\rm 1},
    Harshvardhan C. Takawale\textsuperscript{\rm 1},
    Yehia Zakaria\textsuperscript{\rm 2},
    Ahmed Allam\textsuperscript{\rm 2},
    Nirupam Roy\textsuperscript{\rm 1}
}
\affiliations {
    \textsuperscript{\rm 1}Department of Computer Science, University of Maryland, College Park\\
    \textsuperscript{\rm 2}Department of Mechanical and Materials Engineering, University of Cincinnati\\
    \{ssadhu1, htakawal, niruroy\}@umd.edu, \{allamad, saidyz\}@ucmail.uc.edu
}

\begin{document}

\maketitle

\begin{abstract}
Recent work in additive manufacturing and neural inverse design enables the creation of complex three-dimensional structures to manipulate physical fields, yet these designs must also adhere to discrete, non-differentiable fabrication constraints. Current methods limit the design space by imposing differentiable soft manufacturability loss penalties, or by fixing structures through post-processing, causing a discrepancy between the geometry optimized by the physics model and the final fabricated result. Thus, we present \sysName, a physics-informed neural architecture for constrained inverse design that represents three-dimensional geometries with an implicit neural representation (INR) and passes each candidate through a printability operator before evaluating its final response. The printability operator executes layer-wise self-support, discrete material selection, and whole-body connectivity, assuring fabrication feasibility. To allow for backpropagation through the discrete non-differentiable operations, we use surrogate gradients for the backward pass and a self-supervised printability objective that progressively drives the INR toward a valid design domain. We demonstrate \sysName\ for personalized transcranial focused ultrasound, where inter-patient differences in acoustic properties and head geometry deform wave propagation and requires patient-specific acoustic focusing. \sysName\ jointly optimizes three-dimensional geometry and material composition through differentiable wave physics while ensuring manufacturability. Experiments show manufacturable acoustic structures with enhanced focal accuracy and over density-based and phase-based baselines, demonstrating a generalized methodology for neural design through non-differentiable manufacturing constraints and differentiable physics.

\end{abstract}


\section{Introduction}
Recent advances in additive manufacturing have dramatically expanded the design space of physical systems, enabling complex volumetric geometries and spatial material compositions that were previously difficult or impossible to fabricate. In parallel, neural representations coupled with differentiable physics have emerged as a powerful paradigm for inverse design, allowing structures to be optimized directly for desired physical responses. Such approaches can design structures that manipulate physical fields—from electromagnetic structures for antennas and photonics to acoustic structures that shape wave fields for sensing and medical intervention. Yet exploiting this expanded design space introduces a fundamental challenge: structures that optimize a physical objective are not necessarily realizable. Fabrication imposes discrete material choices, layer-wise support requirements, structural connectivity, and other constraints that are inherently non-differentiable. Existing approaches typically restrict the design parametrization, incorporate manufacturability as soft penalties, or repair unconstrained designs through post-processing. These strategies either sacrifice the expressive volumetric design space or create a mismatch between the geometry evaluated by the physics model and the geometry ultimately fabricated. The resulting challenge is to retain the expressiveness of neural inverse design while guaranteeing fabrication feasibility throughout physics-driven optimization.

We introduce \sysName, a physics-informed neural framework for inverse design under discrete fabrication constraints. Our central principle is to treat manufacturability as an operator on the neural design itself, rather than as a penalty competing with the physical objective. An implicit neural representation provides an expressive continuous design space, while a printer-parameterized feasibility operator maps each candidate design into the realizable set before evaluation by the differentiable physics solver. Thus, optimization is always driven by the physical response of a structure that can actually be fabricated. To reconcile exact feasibility with gradient-based learning, \sysName\ decouples forward and backward computation: discrete operations enforce material occupancy and global structural connectivity in the forward pass, while structurally matched surrogate gradients provide informative learning signals through their non-differentiable components. A self-supervised printability objective further drives the neural representation toward the feasible set, reducing the discrepancy between the generated and realizable geometries over training.

\noindent {\bf Representative application.} We instantiate \sysName\ for personalized transcranial focused ultrasound using a structure, called acoustic lens, which focuses acoustic energy to localized brain regions for procedures including neuromodulation, blood-brain-barrier opening, and tumor treatment. Precise focusing is challenging because the skull strongly distorts ultrasound propagation, while its geometry, tissue composition, and acoustic properties vary across individuals. We quantify this variability using 290 head CT scans from the Qure.ai dataset, converting each scan into heterogeneous sound-speed, density, and attenuation fields. Across patients, cortical-skull sound speed alone can vary about 30\%;
%, while cortical-skull composition ranges from 6.49\% to 19.18\%; 
pairwise analysis further reveals substantial variation in skull surface geometry and volume that impacts the acoustic field. More importantly, due to differences in acoustic medium, a lens across patients may lead to focal point displacements of 7.67–21.99 mm, demonstrating why millimeter-scale targeting requires patient-specific design. \sysName\ turns this personalization problem into an automated neural inverse-design pipeline: given a patient's CT-derived acoustic model and desired focal region, it jointly optimizes the volumetric geometry and material composition of an acoustic structure through differentiable wave physics while ensuring that the resulting design remains directly manufacturable.

Our contributions in this paper are as follows:
\begin{itemize}
    \item {\bf Constrained neural inverse design:}
    %We formulate manufacturable physical design as optimization through a printer-parameterized feasibility operator, enforcing fabrication constraints by construction during physics-driven learning.
    Optimized manufacturable physical designs through a printer-parameterized feasibility operator that enforces fabrication constraints by construction.
    \item {\bf Differentiable printability:}
    %We incorporate discrete material selection, recursive layer-wise self-support, and global structural connectivity, with structurally matched surrogate gradients for non-differentiable operations.
    Incorporated discrete material selection, layer-wise self-support, and structural connectivity using structurally matched surrogate gradients.
    \item {\bf Personalized acoustic geometry synthesis:} 
    %We instantiate the framework for transcranial focused ultrasound and demonstrate directly printable volumetric lenses with improved focal accuracy and acoustic energy concentration over phase- and density-based baselines.
    Instantiated the framework for transcranial focused ultrasound, producing directly printable volumetric lenses with improved focal accuracy and acoustic energy concentration over phase- and density-based baselines.
\end{itemize}    

\section{Related Works}
\noindent {\bf Differentiable Physics and Neural Inverse Design. }
Differentiable simulators~\cite{zhu2021general,stanziola2023j,liu2025deep} have allowed for simulation-in-the-loop optimization, where numerical solvers can be placed into the gradient-based inverse design pipeline\cite{lee2024machine}. This capability has accelerated inverse design across a wide range of physical systems, such as acoustics~\cite{fushimi2021acoustic}, electromagnetics,~\cite{minkov2020inverse} and mechanics~\cite{pastrana2025real}, in which complex geometries can be optimized directly with respect to physical objectives. However, most differentiable inverse design approaches assume optimized geometry can be manufactured exactly, overlooking constraints imposed by additive manufacturing.

\noindent {\bf Neural Design for Additive Manufacturing. }
Recently, there has been growing interest in embedding neural networks directly into the topology optimization pipeline for additive manufacturing (AM). AM design has been framed as a generative modeling problem to produce diverse design candidates~\cite{nie2021topologygan,yamasaki2021data, giannone2023aligning}. Additionally, manufacturing constraints such as self-support have been applied via loss regularizers~\cite{chen2024atlas3d,dang2026densicrafter} as well as differentiable 3D structure slicing formulations that jointly optimize manufacturable layer generation and printing trajectories~\cite{liu2024neural,liu2025neural}. Although these methods demonstrate that fabrication constraints can be easily integrated into the gradient optimization pipeline, they primarily optimize structural performance, such as maximizing stiffness, minimizing compliance, or ensuring printability. In contrast, we co-optimize a printable acoustic geometry for wave focusing, specifically demonstrating focusing efficacy in the transcranial ultrasound application. 

\noindent {\bf Acoustic Inverse Design for Transcranial Ultrasound. }
Acoustic lenses are commonly designed to compensate for wave distortions by introducing phase delays, which is geometrically represented via thickness-based lenses~\cite{melde2016holograms,maimbourg20183d,jimenez2019holograms}. However, this limits the design space (cannot exploit internal volumetric structures) and assumes a single material. Density-based optimization~\cite{lee2024machine,stanziola2023physics,hyun2021transient} relaxes this assumption by optimizing a continuous density field; but, it does not explicitly enforce manufacturability, while some methods rely on post-processing or hard thresholding to obtain printable geometries~\cite{stanziola2023physics}. Unlike prior work, we jointly optimize volumetric acoustic performance and manufacturability, embedding a differentiable printability operator into the inverse design pipeline.

\begin{figure*}[t]
    \centering
    \includegraphics[width=\textwidth]{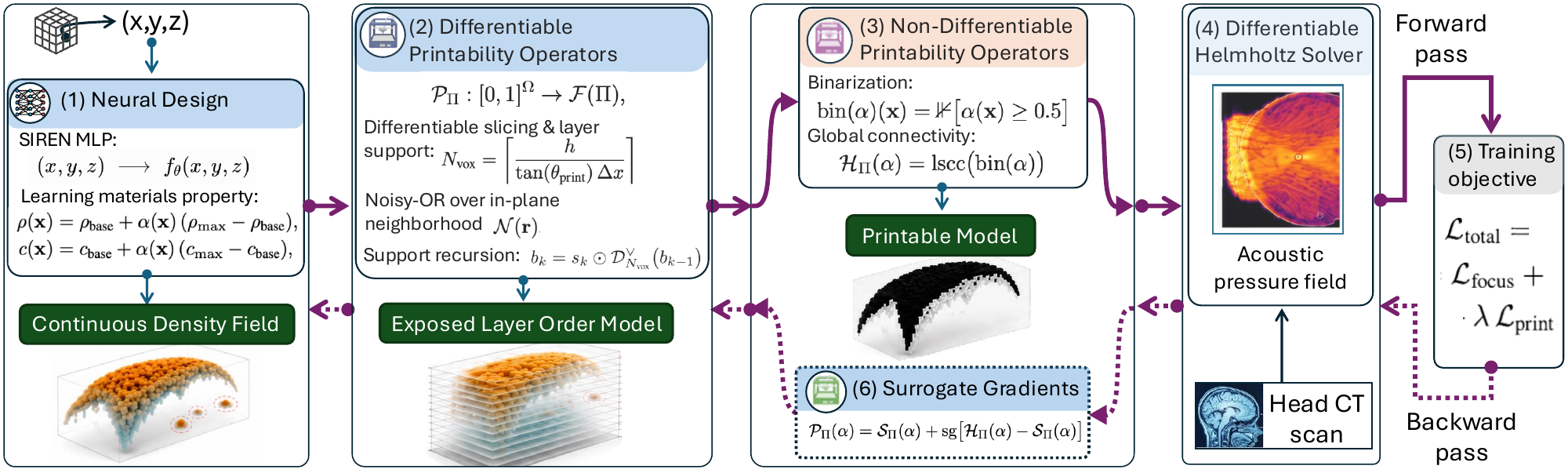}
    %{Figures/Methods/Lens Optimization.pdf}
    % \caption{Pipeline for \sysName. We optimize neural acoustic geometries with a combination of custom differentiable and non-differentiable components. }
    \caption{Overview of \sysName. End-to-end optimization of printable acoustic geometries.}
    \label{fig:Methodspipeline}
\end{figure*}

\section{Method}
\label{sec:method}

We describe personalized acoustic geometry design as \emph{inverse design of a heterogeneous material geometry under fabrication constraints}. The central idea is that manufacturability should not be a penalty term appended to a physics objective~\cite{chen2024atlas3d,dang2026densicrafter}, nor a restriction baked into the parametrization, but a \emph{differentiable operator applied to the design itself}, allowing the wave solver to observe a printable geometry at every optimization step. In Problem Formulation, we formalize the problem and the physics. Neural Design Parametrization introduces the neural parametrization. Differentiable Printability constructs the printability operator and Surrogate Backpropagation develops the surrogate gradients that make its non-differentiable components trainable.

\subsection{Problem Formulation}
\label{sec:problem}

Let $\Omega \subset \mathbb{R}^3$ be the simulation domain, discretized on a
uniform grid with spacing $\Delta x$. A background medium
$\mathcal{M} = \big(c_0(\mathbf{x}),\, \rho_0(\mathbf{x}),\,
a_0(\mathbf{x})\big)$ specifies the spatially varying sound speed, density,
and attenuation of everything that is not the geometry. In our experiments
$\mathcal{M}$ is derived per patient from a head CT scan, but the
formulation is agnostic to its origin and applies to any heterogeneous
medium.

A printer specification $\Pi = \big(\mathcal{P},\,
\theta_{\text{print}},\, h,\, \hat{\mathbf{e}}\big)$ collects the
printable material set $\mathcal{P} = \{(\rho_i, c_i)\}_{i=1}^{K}$, the
maximum self-supporting overhang angle $\theta_{\text{print}}$, the layer
height $h$, and the build direction $\hat{\mathbf{e}}$. Together these
induce a \emph{feasible set} $\mathcal{F}(\Pi)$ of geometries realizable on
that printer without sacrificial support structure.

Given a target focal region encoded by a mask $m_{\text{focus}}$ and a fixed
source configuration, we seek an acoustic geometry design field $\alpha : \Omega \to
[0,1]$ solving
\begin{equation}
    \max_{\alpha}\;
    \mathcal{J}\big(p[\alpha, \mathcal{M}]\big)
    \quad \text{subject to} \quad
    \alpha \in \mathcal{F}(\Pi),
    \label{eq:constrained}
\end{equation}
where $p[\alpha, \mathcal{M}]$ is the steady-state pressure field obtained by solving the Helmholtz equation for the combined geometry-plus-medium
configuration, and $\mathcal{J}$ scores focal quality.

Equation~\eqref{eq:constrained} is a constrained optimization over a
discrete, non-convex feasible set: $\mathcal{F}(\Pi)$ is defined by
binary material occupancy, layer-wise support propagation along
$\hat{\mathbf{e}}$, and global structural connectivity to the build plate.
Rather than relax these constraints into soft penalties, we construct a differentiable operator
\begin{equation}
    \mathcal{P}_{\Pi} : [0,1]^{\Omega} \to \mathcal{F}(\Pi),
    \label{eq:operator}
\end{equation}
that maps an arbitrary design field onto a printable one, and optimize the
unconstrained composition
\begin{equation}
    \max_{\theta}\;
    \mathcal{J}\big(p[\,\mathcal{P}_{\Pi}(\alpha_\theta),\, \mathcal{M}\,]\big).
    \label{eq:composed}
\end{equation}
Feasibility therefore holds \emph{by construction} at every iteration, and
the simulator is never exposed to a geometry that could not be built.

\paragraph{Differentiable Helmholtz Solver}
\label{sub:prelim}
 
We evaluate $p[\alpha, \mathcal{M}]$ with j-Wave~\cite{stanziola2023j}, a
differentiable acoustic simulator built on JAX~\cite{jax2018github}. We
use its frequency-domain solver, which returns the steady-state field
directly without time stepping; this suffices because focused ultrasound is
driven continuously at a single frequency. The governing time-harmonic
equation, derived from the second-order wave equation with Stokes
absorption, is
\begin{equation}
    -\frac{\omega^2}{c_0^2}\,P
    =
    \nabla^2 P
    - \frac{1}{\rho_0}\,\nabla\rho_0 \cdot \nabla P
    + \frac{2i\omega^3 a_0}{c_0}\,P
    - i\omega S_M,
    \label{eq:helmholtz}
\end{equation}
where $S_M$ is the source term and $c_0$, $\rho_0$, $a_0$ are the sound
speed, density, and attenuation at each point. The acoustic geometry enters through
precisely these coefficients: it occupies a prescribed region of $\Omega$
and sets the material properties within it, so designing an acoustic geometry amounts to
choosing the spatially varying coefficients of
Eq.~\eqref{eq:helmholtz}. Discretized on a grid, the equation becomes a large sparse linear
system, which j-Wave solves with a matrix-free iterative method. Gradients
with respect to the material parameters are obtained by implicit
differentiation rather than by backpropagating through every solver
iteration. j-Wave is therefore fully differentiable and can be inserted
directly into the gradient-based optimization pipeline developed below. j-Wave has been validated against a cohort of
transcranial ultrasound simulators~\cite{aubry2022benchmark}.

\subsection{Neural Design Parametrization}
\label{sec:inr}

\paragraph{Implicit design field}
We parametrize the acoustic geometry with a SIREN MLP~\cite{sitzmann2020implicit}
$f_\theta$, chosen for its ability to represent high-frequency spatial
detail without the spectral bias of ReLU coordinate networks. The network
does not predict acoustic properties directly, it emits a scalar
\emph{occupancy} logit per coordinate, whose activation
$\alpha(\mathbf{x}) \in [0,1]$ sets the density $\rho(\mathbf{x})$ and sound
speed $c(\mathbf{x})$ at each voxel by interpolating between the base medium
and the acoustic geometry material:
\begin{align}
    \rho(\mathbf{x}) &= \rho_{\text{base}} + \alpha(\mathbf{x})\,(\rho_{\max} - \rho_{\text{base}}), \label{eq:rho_interp}\\
    c(\mathbf{x})    &= c_{\text{base}}    + \alpha(\mathbf{x})\,(c_{\max}    - c_{\text{base}}), \label{eq:c_interp}
\end{align}
where $(\rho_{\text{base}}, c_{\text{base}})$ are the density and sound speed
of water ($1000\,\mathrm{kg/m^3}$, $1480\,\mathrm{m/s}$) and
$(\rho_{\max}, c_{\max})$ bound the achievable material properties. Representing the acoustic geometry implicitly
decouples design resolution from network capacity and provides a smooth,
mesh-free geometry that the printability operator can act on directly.

\paragraph{Focusing objective}
We maximize acoustic energy delivered to the target region rather than to a
single voxel, which avoids degenerate single-point solutions and makes the
objective well-defined for the extended and non-convex focal patterns
evaluated in Figure~\ref{fig:threepatterns}. Writing $\mathcal{J}$ as a loss,
\begin{equation}
\mathcal{L}_{\mathrm{focus}}
=
-\sqrt{
\frac{
\sum_{\mathbf{x}} |p(\mathbf{x})|^{2}\, m_{\mathrm{focus}}(\mathbf{x})
}{
\sum_{\mathbf{x}} m_{\mathrm{focus}}(\mathbf{x}) + \varepsilon
}
},
\label{eq:focus}
\end{equation}
where $p(\mathbf{x})$ is the complex pressure, $m_{\mathrm{focus}}$ is a
binary mask over the desired focal region, and $\varepsilon > 0$ ensures
numerical stability. Minimizing $\mathcal{L}_{\mathrm{focus}}$ maximizes the
RMS pressure inside the target.

\paragraph{Unconstrained reference}
Optimizing Eq.~\eqref{eq:focus} over the continuous field $\alpha$ alone,
with no fabrication constraints, yields what we call the
\emph{unconstrained reference geometry}. It is not physically realizable. It
assumes a continuum of materials and admits floating, disconnected
structure. But it characterizes the focusing performance attainable if
manufacturability were free, and serves as the empirical upper bound against
which we measure the cost of printability.

\subsection{Differentiable Printability Operators}
\label{sec:printop}

We now construct $\mathcal{P}_{\Pi}$ of Eq.~\eqref{eq:operator} as a
composition of four stages: material discretization, slicing, layer-wise
support propagation, and global connectivity extraction. Stages are applied
along the build direction $\hat{\mathbf{e}}$ and parametrized entirely by
$\Pi$, so the same operator instantiates different printers by substituting
their specification. We emphasize that this covers layered additive
processes with a fixed build direction; non-planar and multi-axis
deposition~\cite{liu2024neural} lies outside the present formulation.

\paragraph{Material Discretization}
\label{sec:gumbel}

Physical printers deposit a discrete set of materials, so the continuous
$\alpha$ of Eq.~\eqref{eq:rho_interp} must be quantized. We use a
Gumbel-Softmax activation function~\cite{jang2017categorical,maddison2016concrete},
which allows gradients to flow through discrete sampling operations.

For a two-material printer (acoustic geometry material and air/water, as on FDM or SLA
systems), let $\ell(\mathbf{x}) = f_\theta(\mathbf{x})$ denote the raw
pre-activation output. With $u \sim \mathcal{U}(0,1)$ and logistic noise
$L = \log u - \log(1-u)$, the Binary Concrete relaxation gives
\begin{equation}
    \alpha_{\tau}(\mathbf{x}) = \sigma\!\left(\frac{\ell(\mathbf{x}) + L}{\tau}\right),
    \label{eq:gumbel_sigmoid}
\end{equation}
where $\sigma(\cdot)$ is the logistic sigmoid and $\tau$ is the temperature.
As $\tau \to 0$, $\alpha_\tau$ approaches a Bernoulli sample.

For multi-material printers such as material jetting, discrete selection
over $K$ materials uses the Gumbel-Softmax distribution
\begin{equation}
    y_i = \frac{\exp\!\big((\ell_i + g_i)/\tau\big)}
               {\sum_{j=1}^{K} \exp\!\big((\ell_j + g_j)/\tau\big)},
    \qquad g_i \sim \text{Gumbel}(0,1),
    \label{eq:gumbel_softmax}
\end{equation}
with acoustic properties recovered as the convex combination
\begin{equation}
    \rho(\mathbf{x}) = \sum_{i=1}^{K} y_i(\mathbf{x})\,\rho_i,
    \qquad
    c(\mathbf{x}) = \sum_{i=1}^{K} y_i(\mathbf{x})\,c_i .
    \label{eq:multimat}
\end{equation}
Multi-material capability matters acoustically, not just mechanically: it
lets the optimizer tailor \emph{impedance} throughout the acoustic geometry volume rather
than modulating phase through thickness alone, which is the design freedom
that phase-only acoustic structures fundamentally lack.

\paragraph{Differentiable Slicing}
\label{sec:slicer}

In additive manufacturing, a slicing software converts a 3D model into an ordered sequence of printable layers and determines the fabrication path for each layer. Inspired by this process, 
%Following the logic of 3D slicing softwares, 
we treat the quantized geometry as
an occupancy field and traverse it as a stack of $K$ layers perpendicular to
$\hat{\mathbf{e}}$. Let $s_k$ denote the in-plane occupancy of layer $k$,
with $s_0$ resting on the build plate and therefore unconstrained. Slicing
is a reshaping of the design field and is differentiable, its role
is to expose the layer ordering that support propagation requires.

\paragraph{Layer-wise Self-Support}
\label{sec:support}

Because both the internal and external geometry of the structure are relevant for acoustic behavior, the design must be \emph{self-supporting}: any structural support
inserted by slicer software for printing alters its acoustic behavior. A
voxel is printable only if the layer beneath it provides material within the
lateral reach permitted by the overhang angle. For layer height $h$ and grid
spacing $\Delta x$, that reach is
\begin{equation}
    N_{\text{vox}} = \left\lceil \frac{h}{\tan(\theta_{\text{print}})\,\Delta x} \right\rceil
    \label{eq:nvox}
\end{equation}
voxels. The hard rule is a conjunction between the candidate slice and a
dilation of the layer below,
\begin{equation}
    b_k = s_k \wedge \mathrm{OR}_{N_{\text{vox}}}\big(b_{k-1}\big),
    \label{eq:hard_support}
\end{equation}
which is non-differentiable. We replace the disjunction with a smooth
noisy-OR over the in-plane neighborhood $\mathcal{N}(\mathbf{r})$,
\begin{equation}
\begin{split}
\mathcal{D}^{\vee}(q)(\mathbf{r})
&= 1 - \exp\!\left( \sum_{\mathbf{r}' \in \mathcal{N}(\mathbf{r})} \log\big(1 - q(\mathbf{r}')\big) \right) \\
&= 1 - \prod_{\mathbf{r}' \in \mathcal{N}(\mathbf{r})} \big(1 - q(\mathbf{r}')\big),
\end{split}
\label{eq:softor}
\end{equation}
for $q(\mathbf{r}) \in [0,1]$, and apply it $N_{\text{vox}}$ times to obtain
lateral reach matching Eq.~\eqref{eq:nvox}:
\begin{equation}
    q^{(0)} = q, \quad
    q^{(m+1)} = \mathcal{D}^{\vee}\big(q^{(m)}\big), \quad
    \mathcal{D}^{\vee}_{N_{\text{vox}}}(q) = q^{(N_{\text{vox}})}.
    \label{eq:dilate}
\end{equation}
The support recursion then reads
\begin{equation}
    b_0 = s_0, \qquad
    b_k = s_k \odot \mathcal{D}^{\vee}_{N_{\text{vox}}}\big(b_{k-1}\big),
    \label{eq:soft_support}
\end{equation}
for $k = 1, \dots, K-1$, with $\odot$ the elementwise product acting as a
soft conjunction.

Equation~\eqref{eq:soft_support} is the operator's essential mechanism. It
is \emph{recursive}. $b_k$ depends on $b_{k-1}$,
which depends on $b_{k-2}$, the printability of any voxel is a function of
the entire column of material beneath it. A local overhang penalty evaluated
independently at each voxel cannot express this dependency, and consequently
cannot distinguish a genuinely grounded structure from one resting on
material that is itself unsupported. Gradients likewise propagate down the
stack, so the optimizer receives signal about how a change at one layer
affects buildability everywhere above it.

\subsection{Non-Differentiable Printability Operators}
\paragraph{Binarization and Global Connectivity}
\label{sec:connectivity}

Two gaps remain between $b_k$ and a truly printable geometry. First, the
Gumbel relaxation concentrates values near $\{0,1\}$ but does not reach
them, so we threshold
\begin{equation}
    \mathrm{bin}(\alpha)(\mathbf{x}) =
    \mathbb{1}\big[\alpha(\mathbf{x}) \geq 0.5\big].
    \label{eq:binarize}
\end{equation}
Second, Eq.~\eqref{eq:soft_support} enforces \emph{local} support but not
\emph{global} grounding: a cluster of voxels can support itself layer over
layer while remaining disconnected from the build plate, forming a floating
island that no printer can produce. We therefore extract the largest
connected component that is structurally rooted at $k=0$, denoted
$\mathrm{lscc}(\cdot)$.

The composition
\begin{equation}
    \mathcal{H}_{\Pi}(\alpha) = \mathrm{lscc}\big(\mathrm{bin}(\alpha)\big)
    \label{eq:hard_op}
\end{equation}
is the exact printability operator: its output is genuinely manufacturable,
and it is exactly what we pass to the wave solver. It is also
non-differentiable in both stages --- thresholding has zero gradient almost
everywhere, and connected-component labeling is a combinatorial graph
operation with no meaningful derivative. The Surrogate Backpropagation Section
resolves this.

\subsection{Surrogate Backpropagation}
%\subsection{Surrogate Gradients for Non-Differentiable Operations}
\label{sec:surrogate}

Due to the presence of non-differentiable operations, a stop gradient must be used to decouple the forward and backward computations. Let
$\mathcal{S}_{\Pi}$ be a differentiable surrogate for $\mathcal{H}_{\Pi}$
and let $\mathrm{sg}[\cdot]$ denote the stop-gradient operator. The
printability operator of Eq.~\eqref{eq:operator} is implemented as
\begin{equation}
    \mathcal{P}_{\Pi}(\alpha)
    = \mathcal{S}_{\Pi}(\alpha)
    + \mathrm{sg}\big[\mathcal{H}_{\Pi}(\alpha) - \mathcal{S}_{\Pi}(\alpha)\big].
    \label{eq:ste}
\end{equation}
In the forward pass Eq.~\eqref{eq:ste} evaluates exactly to
$\mathcal{H}_{\Pi}(\alpha)$, so the simulator sees the true printable
geometry and $\mathcal{L}_{\text{focus}}$ reflects the performance of a lens
that can actually be built. In the backward pass the gradient is
$\partial \mathcal{S}_{\Pi} / \partial \alpha$. 
% This follows the
% straight-through construction of~\cite{bengio2013estimating} as used
% in~\cite{vandenoord2017neural}.

The difficulty is that $\mathrm{lscc}$ has no natural relaxation: a
temperature-annealed version of connected-component labeling does not exist,
and substituting the identity discards precisely the structural information
we need. Our surrogate is therefore not a softened copy of the same
computation but a \emph{structurally matched approximation}, which is a recursive
scoring scheme that reproduces the global grounding semantics of
$\mathrm{lscc}$ using only local, differentiable operations.

\paragraph{Backward supported-mass scoring}
We first go downward from the top of the stack, accumulating
at each voxel a measure of how much material it carries above it. With
$\zeta_{K} \equiv 0$,
\begin{equation}
    \zeta_k = s_k \odot \big(1 + \mathcal{D}^{\vee}_{N_{\text{vox}}}(\hat{\zeta}_{k+1})\big),
    \qquad
    \hat{\zeta}_{k} = \frac{\zeta_{k}}{\max_{\mathbf{r}} \zeta_{k}(\mathbf{r}) + \varepsilon},
    \label{eq:score}
\end{equation}
for $k = K-1, \dots, 0$, where $\hat{\zeta}$ denotes the score field
normalized to $[0,1]$. The dilation $\mathcal{D}^{\vee}_{N_{\text{vox}}}$ is
the same operator used in the forward support recursion
(Eq.~\eqref{eq:dilate}), so the surrogate inherits the lateral reach imposed
by the printer's overhang angle. A voxel accumulates high $\zeta$ when it anchors a tall, well-connected column (the property that
distinguishes the main body from an island) obtained without graph traversal.

\paragraph{Root gating}
At the build plate many candidate roots survive, most of them spurious. We
retain the strongest with a soft spatial gate,
\begin{equation}
    \gamma = \sigma\!\left(\frac{\zeta_0 - \eta \max(\zeta_0)}{\tau_{g}}\right),
    \qquad
    \tilde{b}_0 = s_0 \odot \gamma,
    \label{eq:gate}
\end{equation}
where $\eta \in (0,1)$ sets the relative retention threshold ($\eta = 0.5$
in our experiments) and $\tau_{g}$ controls sharpness. Because $\gamma$ is
sigmoidal rather than a hard selection, gradients flow to \emph{all}
candidate roots in proportion to their support mass, allowing the optimizer
to strengthen a marginal root instead of receiving zero signal once it falls below a threshold.

\paragraph{Upward reconstruction}
Reapplying the support rule from the gated base recovers the surrogate main
body,
\begin{equation}
    \tilde{b}_k = s_k \odot \mathcal{D}^{\vee}_{N_{\text{vox}}}\big(\tilde{b}_{k-1}\big),
    \qquad
    \mathcal{S}_{\Pi}(\alpha) = \tilde{b}_{0:K-1}.
    \label{eq:surrogate_out}
\end{equation}
The result is a differentiable field that agrees closely with
$\mathcal{H}_{\Pi}(\alpha)$ in value while remaining spatially
informative in gradient: each voxel receives signal scaled by its
structural role, rather than the uniform pass of a naive
straight-through estimator. Figure~\ref{fig:ablation} isolates this choice
against straight-through and fully-relaxed alternatives.

\paragraph{Self-Supervision}
\label{sec:selfsup}
 
Gradient signal through Eq.~\eqref{eq:ste} shapes the design only indirectly,
via its effect on the focal field. We additionally use the printable lens as
a direct supervision signal for the continuous field it was derived from,
penalizing the discrepancy between the two:
\begin{equation}
    \mathcal{L}_{\text{print}}
    = \frac{1}{|\Omega|} \sum_{\mathbf{x} \in \Omega}
      \Big(\alpha(\mathbf{x}) - \mathrm{sg}\big[\mathcal{P}_{\Pi}(\alpha)(\mathbf{x})\big]\Big)^{2},
    \label{eq:selfsup}
\end{equation}
where $\alpha$ is the raw INR output and $\mathcal{P}_{\Pi}(\alpha)$ its
printable counterpart. This is similar to a self-distillation, or
teacher--student, arrangement, in which the printable lens supervises the
continuous field it was derived from. Rather than relying on gradients
through $\mathcal{P}_{\Pi}$ alone to discourage unprintable structure, the
INR is taught to emit fields that already look printable, while still
backpropagating through a soft version of the lens. As training proceeds, the
projection in Eq.~\eqref{eq:ste} therefore removes progressively less
material, and the discrepancy the surrogate must bridge shrinks. This is empirically shown in Figure~\ref{fig:ablation} that as training progresses voxel removal fraction decreases, showing the INR learning to create a more self-supporting structure.

\subsection{Training Objective}
\label{sec:objective}

The complete objective co-optimizes focusing and manufacturability:
\begin{equation}
    \mathcal{L}_{\text{total}}
    = \mathcal{L}_{\text{focus}}
    + \lambda\, \mathcal{L}_{\text{print}},
    \label{eq:total}
\end{equation}
where $\lambda$ balances the two terms. We stress that $\lambda$ does not
control whether the design is printable. The feasibility is guaranteed by $\mathcal{P}_{\Pi}$ regardless of $\lambda$. It only determines how strongly the continuous field is pulled toward the feasible set, and therefore how much structure the projection must discard at each step. This is the practical distinction between our formulation and penalty-based approaches, in which the constraint weight trades off directly against feasibility and printability is only achieved in the limit.

\section{Evaluation}
\label{sec:setup}

\paragraph{Data.}
We use 290 non-contrast head CT scans from the Qure.ai
dataset~\cite{chilamkurthy2018development}, retaining every scan with at
least 100 axial slices, and evaluating on the three most acoustically varying models. Scans are converted to acoustic property fields
$(\rho, c, a_0)$ using established Hounsfield-unit
mappings~\cite{schneider1996calibration,mast2000empirical} with
tissue-specific attenuation
constants~\cite{pinton2012attenuation,hoffmeister2025fundamental,slominski2023influence}.
Appendix gives the full conversion. 
\paragraph{Experimental Settings.}
We trained on an NVIDIA A6000 for 30 steps with Adam optimizer($lr=5e^{-4}$) and global seed of 42. For hyperparameters, we used $\omega=90$, $\tau=0.05$, $\tau_g=0.01$, and $\lambda=0.02$. Ablation available in Appendix.

\paragraph{Simulation.}
Domains are resampled to $0.4\,$mm, giving $8.1$ points per wavelength. Unless stated
otherwise, the source is a $10 \times 10$ grid of radial transducers driven
at $500\,$kHz and embedded in the first lens layer, and the lens covers
$60\%$ of the head surface. Appendix details the domain
configuration. 
\paragraph{Baselines.}
We compare against four phase-only and density-based alternatives.
\emph{Homogeneous PAT} applies geometric phase delays computed under a
water-only assumption, isolating the cost of ignoring aberration entirely.
\emph{Heterogeneous PAT}~\cite{aubry2003experimental,marquet2009non}
, our strongest phase-only comparison, simulates each array element through the patient's CT-derived medium to
obtain per-element transfer coefficients, then applies phase conjugation.
\emph{Thickness-based lens}~\cite{jimenez2019holograms,hu2022binary}
converts a required phase map into a wrapped single-material thickness
profile, an example of a phase-only lens.
\emph{Sigmoid lens}~\cite{stanziola2023physics,lee2024machine} optimizes a
continuous density field through a sigmoid activation, representing
density-based topology optimization without fabrication constraints.
Formulations in Appendix.

\paragraph{Metrics.}
We report five quantities over the focal region. \emph{Overlap} is the
fraction of the target region exceeding $50\%$ of peak pressure, measuring
how well an anatomical target is covered rather than a single
point. \emph{Ratio} (R/2R) is the ratio of RMS pressure inside the focus to that in a concentric region of twice the radius, measuring concentration against leakage. \emph{Shift} is the distance between the realized and
intended focus centers. \emph{Peak} and \emph{RMS pressure} are reported in
arbitrary units, as j-Wave does not output calibrated pressure.
 
\subsection{Results}

\begin{table}[t]
\centering
\small
\resizebox{\columnwidth}{!}{% 
\begin{tabular}{@{} l *{5}{c} @{}} 
\hline
\textbf{Method} & 
\begin{tabular}{@{}c@{}}\textbf{Overlap} \\ \textbf{(\%)}\end{tabular} & 
\begin{tabular}{@{}c@{}}\textbf{Ratio} \\ \textbf{(R/2R)}\end{tabular} & 
\begin{tabular}{@{}c@{}}\textbf{Shift} \\ \textbf{(mm)}\end{tabular} & 
\begin{tabular}{@{}c@{}}\textbf{Peak P.} \\ \textbf{(A.U)}\end{tabular} & 
\begin{tabular}{@{}c@{}}\textbf{RMS P.} \\ \textbf{(A.U)}\end{tabular} \\ \hline
Unconstrained Ref. & 32.5 & \textbf{1.848} & 0.98 & \textbf{2.34} & \textbf{0.763} \\
\sysName & \textbf{46.6} & 1.808 & \textbf{0.81} & 1.87 & 0.680 \\
Homo. PAT & 32.3 & 1.556 & 4.41 & 0.626 & 0.199 \\
Hetero. PAT & 22.2 & 1.795 & 0.90 & 1.05 & 0.416 \\
Thickness & 24.1 & 1.130 & 4.28 & 0.285 & 0.076 \\
Sigmoid & 27.5 & 1.615 & 1.82 & 0.867 & 0.276 \\ \hline
\end{tabular}%
}
\caption{Main Results for Center Focusing. Average of CT-105, CT-166, and CT-260 results.}
\label{tab:results_center}
\end{table}

Table~\ref{tab:results_center} demonstrates that for a sphere of radius 4 mm, both the unconstrained reference and \sysName\ lens outperform baselines in all five metrics for center focusing. Experiment configuration details in the Appendix.

%We observe a similar pattern when we try off-center focusing as shown in Table~\ref{tab:results_offcenter}, where our method outperforms all the baseline across all metric except the shift metri where Heterogeneous PAT performs better. \todoharsh{Explain why?} 
% \begin{figure*}[t]
%     \centering
%     \includegraphics[width=\textwidth]{Figures/Ablation/ablation.pdf}
%     \caption{Ablation Study of Differentiable and Non-differentiable components across four metrics: (a) Percentage of voxels removed, (b) Largest component fraction, (c) Number of supported components,and (d) RMS Pressure Focus }
%     \label{fig:ablation}
% \end{figure*}

\subsection{Acoustic Lens Comparisons}
% \begin{figure}[H]
%     \centering
%     \includegraphics[width=0.5\columnwidth]{Figures/PressureComparision/beforeafterslicerv2.pdf}
%     \caption{Slicer Comparision between Printable-trained lens and Non-Printable-trained lens. Areas not printable by Ultimaker 3D Slicer by the lens highlighted in \textcolor{orange}{orange}}
%     \label{fig:beforeafter}
% \end{figure}
% A pair of Printable and Non-Printable lenses are passed through Ultimaker Cura where toolpaths are given of each lens. These toolpaths reveal areas of the lens that are unprintable, as shown in Figure~\ref{fig:beforeafter}. When comparing the lenses of the printable and unprintable lenses after being pass through the slicer, the printable lens has a cosine similarity of \textbf{0.99} and \textbf{0.84} for the unprintable lens.

% Additionally, a new pressure field is calculated using the Ultimaker-modified lenses. The difference between the printable and printable-modified pressure fields show a \textbf{0.998} cosine similarity while the non-printable and non-printable modified lenses have a cosine similarity of \textbf{0.83}. Figure~\ref{fig:pressurediff1} illustrates the pressure field difference before and after the 3D slicer. The change in the printable lens shows minute difference compared to the upper-bound lens. 

\begin{figure}[h]
    \centering
    \includegraphics[width=0.95\columnwidth]{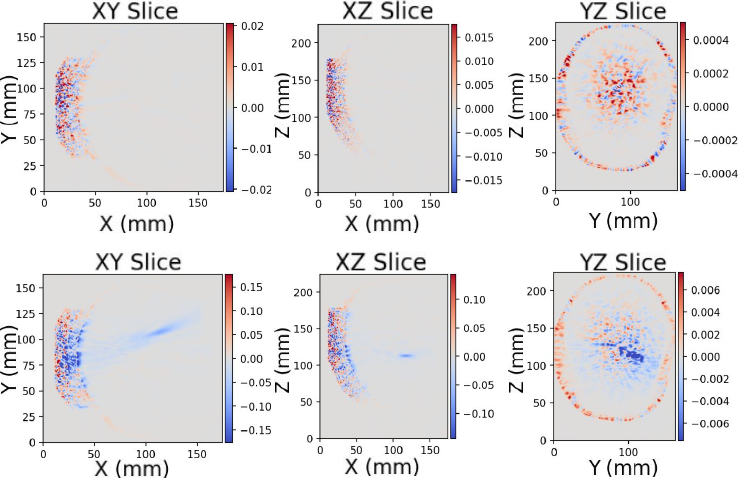}
    \caption{Normalized Pressure Difference of Printable(Top) and Unconstrained(Bottom) lenses before and after 3D Slicer.}
    \label{fig:pressurediff1}
\end{figure}

We validate that our printability constraints hold under a real slicer by passing a printable-trained lens and an unconstrained-trained lens through Ultimaker Cura~\cite{Ultimaker_Cura}, which generates the fabrication toolpaths and highlights areas that cannot be printed as-is. Comparing each lens before and after slicing, the printable lens is almost unchanged (cosine similarity 
0.99), whereas the unconstrained lens is substantially altered (0.84), confirming that the slicer must add or remove material to render it manufacturable. % (Figure~\ref{fig:beforeafter}).
To confirm that these geometric changes are acoustically negligible for our lens, we recompute the pressure field for both slicer-modified geometries. Figure~\ref{fig:pressurediff1} shows that the printable lens's field is essentially preserved (cosine similarity 
0.998 before versus after slicing), while the unconstrained lens's field is markedly degraded (0.83), as the areas of high importance are not printable by the slicer.

\subsection{Printability Ablation Study}
We experiment with the use of differentiable and non-differentiable components for printability. Each pipeline is evaluated on four metrics during training: \emph{RMS Pressure}, \emph{Number of supported connected components}, \emph{Largest supported connected fraction}, and \emph{Percentage of voxels removed}.

The upper-bound and lower-bound pipelines for this study are \emph{Near Printable} and \emph{Printable} respectively. Near Printable consists of all differentiable components, however it cannot be printed due to unsupported voxel islands. Printable performs two non-differentiable operations before the j-Wave simulation, while not using a surrogate function during backpropagation: binarization and largest supported component.

\begin{figure*}[t]
    \centering
    \includegraphics[width=\textwidth]{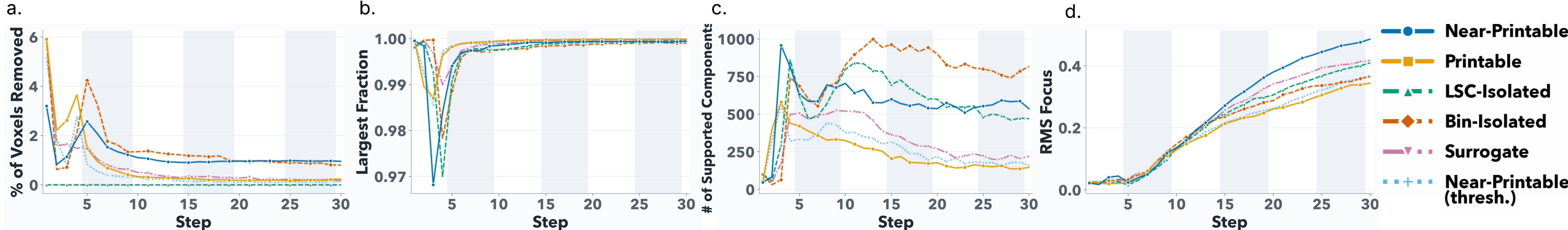}
    \caption{Ablation Study of Differentiable and Non-differentiable components across four metrics: (a) Percentage of voxels removed, (b) Largest component fraction, (c) Number of supported components, and (d) RMS Pressure Focus }
    \label{fig:ablation}
\end{figure*}

\emph{LSC-isolated} and \emph{Bin-isolated} isolate the non-differentiable components of Printable pipeline. Only binarization is non-differentiable in LSC-isolated while largest supported component is non-differentiable for Bin-isolated. \emph{Near Printable(thresholded)} is a printable version of the near printable pipeline where the final lens is thresholded to create a printable lens. The \emph{Surrogate} pipeline was presented in the Methods section.

Figure~\ref{fig:ablation} illustrates that Surrogate has the best RMS of all printable lenses. Additionally, the Surrogate pipeline learns to create fewer supported components, fewer removable voxels, and a larger connected component as effective as the Printable pipeline. A differentiable surrogate component in the backward pass provides the INR with printability guidance to create a printable and better focusing lens.

\subsection{Other Experiments}

\paragraph{Effect of different frequencies}

\begin{table}[h!]
\centering
\small
\setlength{\tabcolsep}{2.5pt} % Reduces padding so all headers fit cleanly
\begin{tabularx}{\columnwidth}{@{} >{\scriptsize\raggedright\arraybackslash}X *{5}{>{\scriptsize\centering\arraybackslash}c} @{}}
\hline
\textbf{Freq(kHz)} & \textbf{Overlap (\%)} & \textbf{Ratio(R/2R)} & \textbf{Shift (mm)} & \textbf{Peak P.(A.U)} & \textbf{RMS P.(A.U)} \\ \hline
100 & 21.3 & 0.691 & 11.22 & 0.4298 & 0.2120 \\
250 & 72.2 & 1.827 & 0.40 & 9.5500 & 4.4780 \\
500 & 26.8 & 1.809 & 0.98 & 3.057 & 0.942 \\
750 & 34.7 & 1.565 & 4.02 & 0.0642 & 0.0217 \\ \hline
\end{tabularx}
\caption{Results for different transducer frequencies.}
\label{tab:results_numfrequency}
\end{table}

Low frequencies(100kHz) result in longer wavelengths and have trouble focusing, however, Table~\ref{tab:results_numfrequency} illustrates that a 250kHz-trained lens is able to perform the best while dealing with the high attenuation of the skull. As frequency increases, absorption of energy also increases resulting in lower Peak and RMS pressure. Visual results are available in Appendix.

\paragraph{Results for different focusing locations}

\begin{table}[h!]
\centering
\small
\resizebox{\columnwidth}{!}{% Scales the table to fit the column perfectly
\begin{tabular}{@{} l *{5}{c} @{}} % Standard tabular with 'l' ensures no line wrapping
\hline
\textbf{Location} & 
\begin{tabular}{@{}c@{}}\textbf{Overlap} \\ \textbf{(\%)}\end{tabular} & 
\begin{tabular}{@{}c@{}}\textbf{Ratio} \\ \textbf{(R/2R)}\end{tabular} & 
\begin{tabular}{@{}c@{}}\textbf{Shift} \\ \textbf{(mm)}\end{tabular} & 
\begin{tabular}{@{}c@{}}\textbf{Peak P.} \\ \textbf{(A.U.)}\end{tabular} & 
\begin{tabular}{@{}c@{}}\textbf{RMS P.} \\ \textbf{(A.U.)}\end{tabular} \\ \hline
Center &  26.8 & 1.809 & 0.98 & 3.057 & 0.942 \\
Off-Center & 42.7 & 1.821 & 0.98 & 1.777 & 0.66 \\
Diagonal Off-Center & 52.3 & 1.758 & 2.83 & 1.165 & 0.4719 \\
Deep Center & 44.3 & 1.865 & 2.83 & 1.905 & 0.7108 \\
Shallow Center & 49.2 & 1.774 & 2.04 & 2.763 & 1.071 \\
Left Lateral & 28.1 & 1.814 & 1.70 & 1.444 & 0.4616 \\
Right Lateral & 32.7 & 1.747 & 1.65 & 1.980 & 0.6514 \\ \hline
\end{tabular}%
}
\caption{Effect of different focus locations.}
\label{tab:results_difflocations}
\end{table}

Table~\ref{tab:results_difflocations} demonstrates that there is an increasing complexity of ultrasound propagation through the head as the target moves away from the geometric center. Since central targets require a symmetric propagation path, the lens can compensate effectively for skull-induced phase distortions. More peripheral locations require acoustic rays to face greater anatomical variability. Visual examples in the Appendix.

\begin{figure}[H]
    \centering
    \includegraphics[width=0.95\columnwidth]{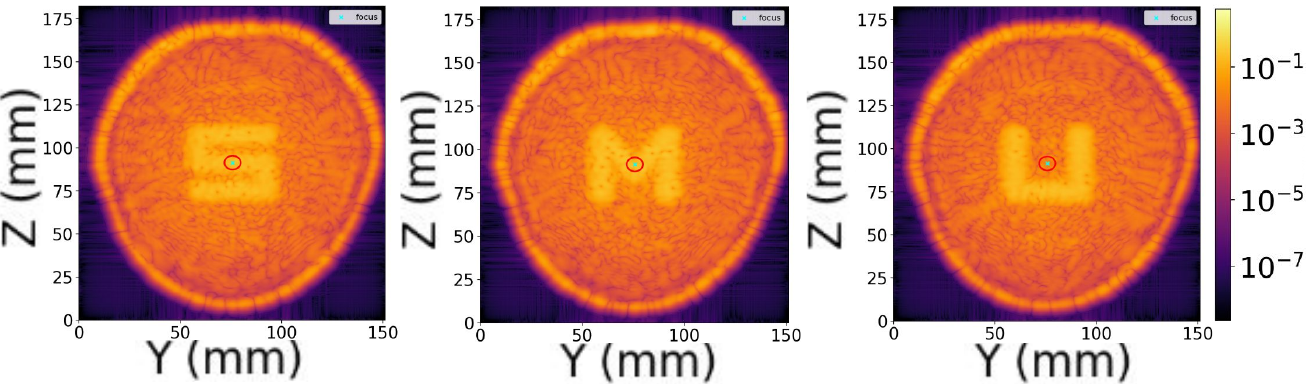}
    \caption{Three letters: `S'(Left), `M'(Middle), `U'(Right).}
    \label{fig:threepatterns}
\end{figure}

\paragraph{Results for shape/pattern matching}
Figure~\ref{fig:threepatterns} demonstrates the acoustic lens's ability to focus on complex patterns. We chose three letters: `M', `U', and `S' These patterns are placed on the YZ slice of the brain with a 5mm thickness, but can also be extended to the other slices(XY and XZ).

\subsection{Cross Simulator Validation}
\begin{figure}[H]
    \centering
    \includegraphics[width=0.95\columnwidth]{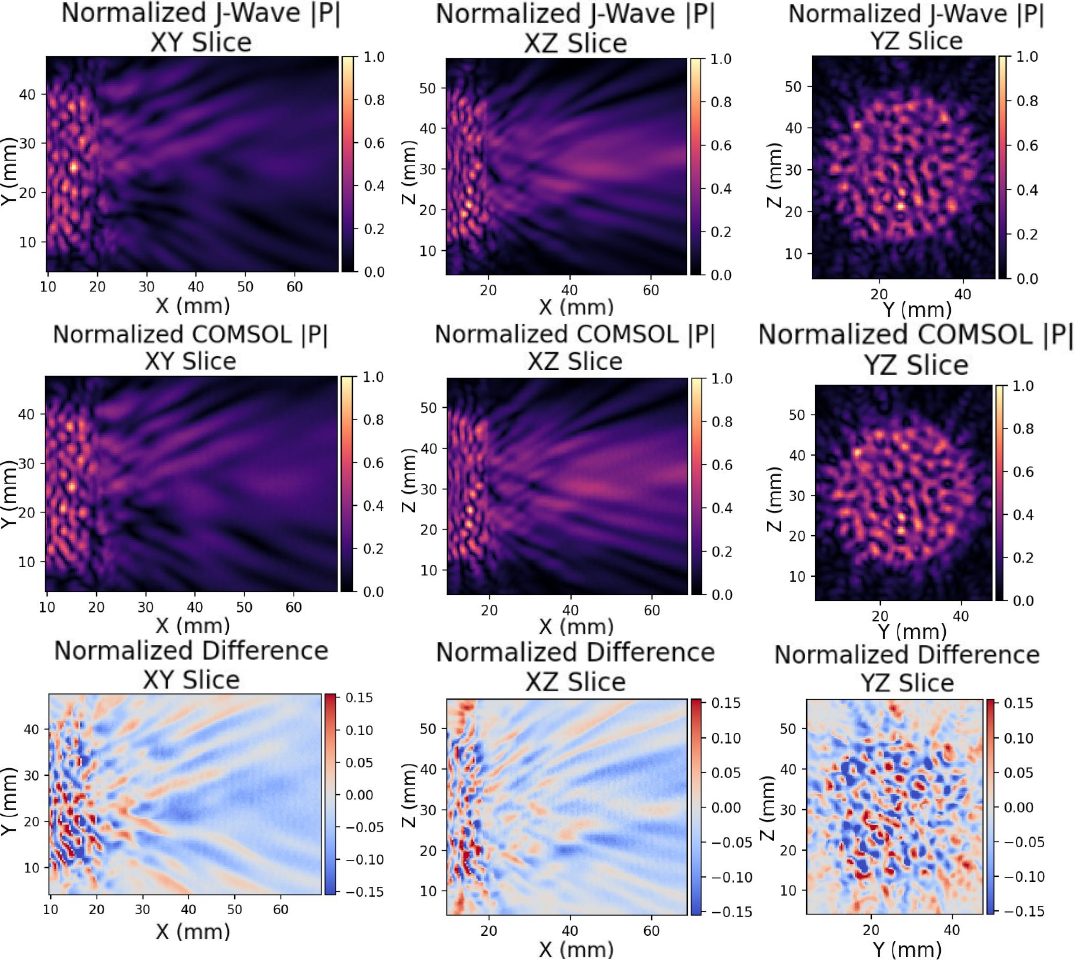}
    \caption{Pressure Field Comparison COMSOL and j-Wave.}
    \label{fig:comsolvjwave}
\end{figure}

Figure~\ref{fig:comsolvjwave} compares the resulting pressure fields of an acoustic lens in j-Wave and COMSOL~\cite{comsol}. Both pressure fields have a cosine similarity of \textbf{0.99}, highlighting the minimal difference between this pair of differentiable and non-differentiable simulators.

\section{Conclusion}
We propose \sysName, a differentiable neural framework for constructing manufacturable acoustic geometries. We design a differentiable slicer that emulates 3D slicers and teaches our implicit geometry to enforce printability. \sysName\ and its unconstrained counterpart demonstrate superior performance to four baselines. We show application in transcranial acoustic focusing, while the core ideas can be extended to other applications.

%\newpage
\clearpage
% \section{Acknowledgments}
% Insert one sentence Acknowledgments.

\bibliography{aaai2027}

\clearpage
\section{Appendix}

\subsection{Analysis of Inter-Patient Variability}
% \begin{figure*}[t]
%     \centering
%     \includegraphics[width=1.0\textwidth]{Figures/Lens Transfer/lenstransferexample.pdf}
%     \caption{Example of lens transfer experiment between CT-105 and CT-11, a structurally similar but acoustically dissimilar CT-scan pair. a) Final pressure field of trained lens on CT-105 for center and off-centered focus and lens transferred onto CT-11. b) Visualization of Iterative Closest Point(ICP) alignment used to find correlated focus point on transferred CT-scan.}
%     \label{fig:transferexample}
% \end{figure*}

Many papers aim to create acoustic lenses that are universal and can be generalized to any patient \cite{lee2024machine,ferri2019enhanced}. Since accurate focusing is critical for delicate organs such as the brain, generalizable lenses can be cause misfocusing that can be critical. Studies measuring skull geometry report that average human skull thickness is approximately 5.97 $\pm$ 1.19 mm, demonstrating substantial variability between individuals \cite{antonakakis2020inter}. Skull thickness can significantly alter the resulting focal field during transcranial ultrasound propagation. Simulation and experimental studies show that variations in skull thickness can change the focal length by approximately 4.4 to 4.7 mm and the axial focal region by 7.6 to 7.9 mm, while also causing more than 80\% attenuation of acoustic intensity in some cases \cite{li2025cross}. One study showed that three human skulls yielded mean sound speeds of 2451 $\pm$ 383 m/s, 2401 $\pm$ 307 m/s, and 2887 $\pm$ 412 m/s \cite{riis2022acoustic}, demonstrating the huge variability in acoustic speed between humans. We reinforce the motivation for personalized acoustic lenses by comparing the CT-scans in our datasets and making 3 arguments: (1) the acoustic properties of each person are different, (2) the structural properties of each person are different, and (3) transferred acoustic lenses between patients can cause at most \textbf{29.59} mm of misfocusing.

\paragraph{Acoustic Differences}
%Increase font size
\begin{figure*}[t]
    \centering
    \includegraphics[width=1.0\textwidth]{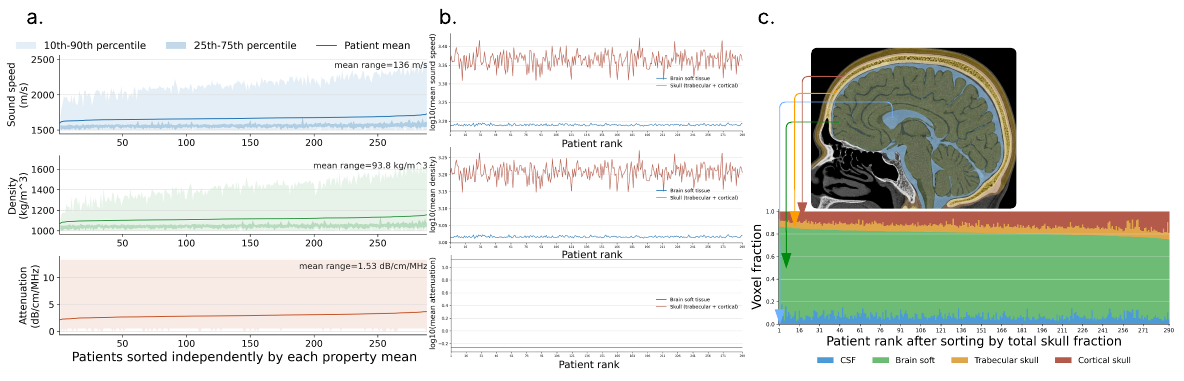}
    \caption{Acoustic Comparison among patients in our dataset for all 290 patients in our dataset. (a) Percentiles of acoustic speed(top), density(middle), and attenuation(bottom). (b) Log-scale of acoustic speed(top), density(middle), and attenuation(bottom) for both brain(blue) and skull(orange) across all patients. (c) Tissue(CSF, Brain, Trabecular Skull, and Cortical Skull) composition.}
    \label{fig:acousticomp}
\end{figure*}

\begin{table}[ht]
\centering
\caption{Acoustic property statistics for the 290 CT-scan dataset}
\label{tab:tissue_statistics}
\resizebox{\columnwidth}{!}{%
\begin{tabular}{llcccc}
\hline
\textbf{Tissue} & \textbf{Parameter} & \textbf{Mean} & \textbf{Std Dev} & \textbf{Min} & \textbf{Max} \\
\hline
Brain Soft Tissue & Speed (m/s) & 1549.6 & 7.1 & 1534.1 & 1569.5 \\
Brain Soft Tissue & Density (kg/m$^3$) & 1039.6 & 5.3 & 1027.9 & 1054.5 \\
Brain Soft Tissue & Attenuation (dB/cm/MHz) & 0.550 & 0.000 & 0.550 & 0.550 \\
\hline
CSF & Speed (m/s) & 1518.7 & 1.1 & 1516.0 & 1523.0 \\
CSF & Density (kg/m$^3$) & 1016.3 & 0.8 & 1014.3 & 1019.5 \\
CSF & Attenuation (dB/cm/MHz) & 0.020 & 0.000 & 0.020 & 0.020 \\
\hline
Trabecular Skull & Speed (m/s) & 1899.2 & 14.2 & 1871.0 & 1953.3 \\
Trabecular Skull & Density (kg/m$^3$) & 1302.4 & 10.7 & 1281.2 & 1343.1 \\
Trabecular Skull & Attenuation (dB/cm/MHz) & 13.300 & 0.000 & 13.300 & 13.300 \\
\hline
Cortical Skull & Speed (m/s) & 2521.1 & 104.9 & 2217.3 & 2857.6 \\
Cortical Skull & Density (kg/m$^3$) & 1770.0 & 78.9 & 1541.6 & 2023.0 \\
Cortical Skull & Attenuation (dB/cm/MHz) & 13.300 & 0.000 & 13.300 & 13.300 \\
\hline
\end{tabular}%
}
\end{table}

After performing our CT conversion to acoustic properties on 290 CT-scans, we were able to perform large-scale data analysis to understand the large acoustic variability among patients. In Table~\ref{tab:tissue_statistics}, the brain soft tissue shows a standard deviation of 7.10 m/s and 5.4 kg/m3 in speed and density respectively. The lack of heterogeneity is due to CT scan's inability to capture brain tissue accurately~\cite{szabo2013diagnostic}. As supported by ~\cite{aubry2003experimental}, Table~\ref{tab:tissue_statistics} displays large variability in speed and density within both parts of the skull: trabecular and cortical. Figure ~\ref{fig:acousticomp}a. shows the variability of speed, density, and attenuation among each patient. Figure ~\ref{fig:acousticomp}b. shows the percentage of tissue composition differs among patients. In Figure~\ref{fig:acousticomp}c, we the log scale variability in speed, density, and attenuation among patients.

\paragraph{Structural Differences}
\label{app:structdiff}

\begin{figure}[H]
    \centering
    \includegraphics[width=1.0\columnwidth]{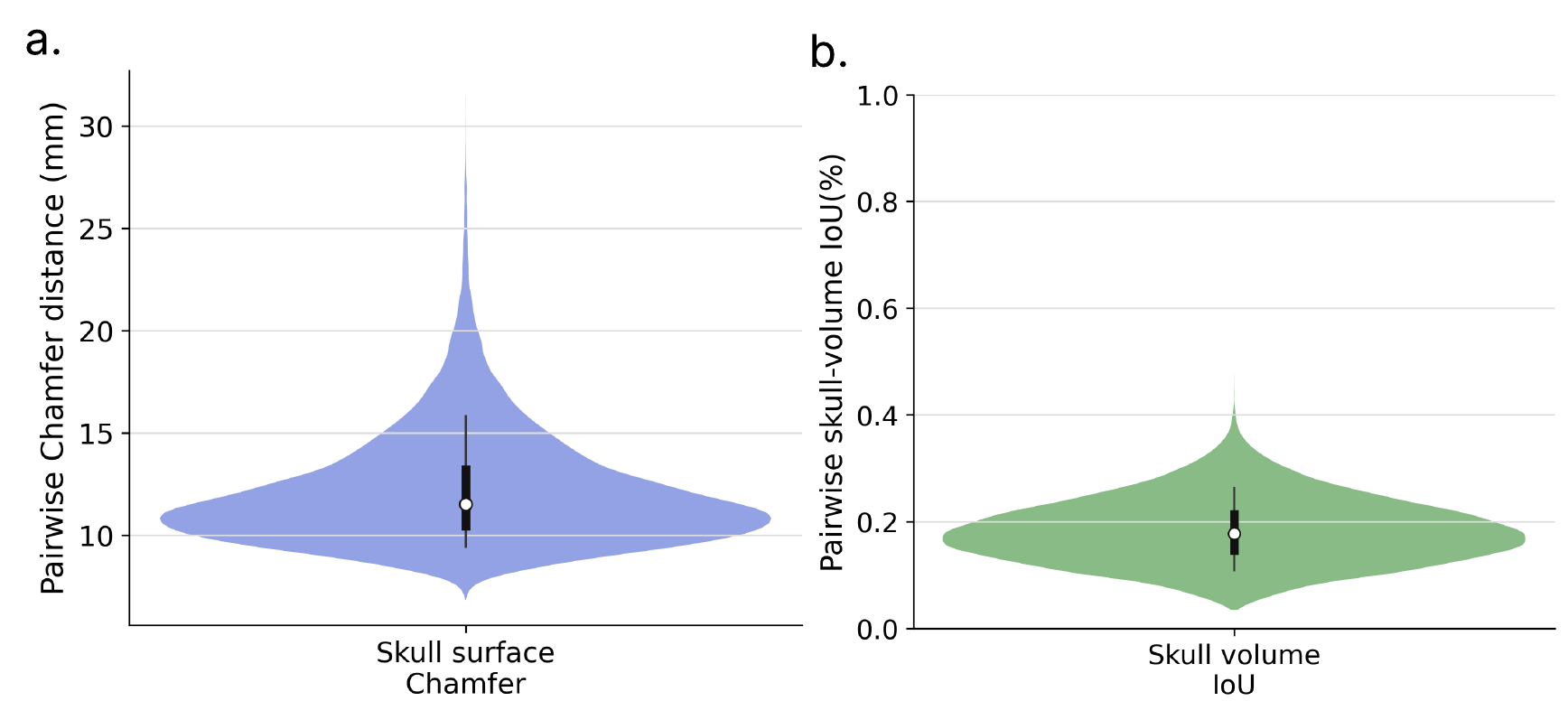}
    \caption{a. Violin Plot of pairwise skull-surface chamfer distances b. Violit Plot of pairwise skull-volume Intersection-over-Union }
    \label{fig:struct_comp}
\end{figure}

\begin{table}[ht]
\centering
\caption{Tissue composition statistics across patients}
\label{tab:tissue_composition}
\resizebox{\columnwidth}{!}{%
\begin{tabular}{lcccc}
\hline
\textbf{Tissue} & \textbf{Mean (\%)} & \textbf{Std Dev (\%)} & \textbf{Min (\%)} & \textbf{Max (\%)} \\
\hline
Brain Soft Tissue & 73.96 & 3.30 & 64.38 & 82.51 \\
\hline
CSF & 6.93 & 2.78 & 2.60 & 17.51 \\
\hline
Trabecular Skull & 6.68 & 1.88 & 3.43 & 15.49 \\
\hline
Cortical Skull & 12.43 & 2.55 & 6.49 & 19.18 \\
\hline
\end{tabular}%
}
\end{table}

Brain soft tissue, Cortical Skull, Trabecular Skull, and CSF can be extracted from CT-scans directly since they all fall in separate Hounsfield ranges. Thus, in Table~\ref{tab:tissue_statistics}, we show that tissue compositions of skull, brain tissue, and CSF all vary among patients. From the CT-scan, we are able to extract body masks of the head and perform structural comparisons between patients. The chamfer distance between patients are calculated to compare head surfaces, while intersection-over-union is calculated to compare head volumes. Figure ~\ref{fig:struct_comp} shows the variation in surface shape and volume among all CT-scans. For both Chamfer Distance and Intersection-Over-Union, we perform Iterative closest point(ICP) alignment on 2 CT-scans. Given patient A and patient B, the rigid transform (rotation + translation) required to algin patient A with patient B is estimated by iterative closest point (ICP). ICP repeatedly matches each point on patient A’s point cloud to the nearest point on B’s cloud, solves for the least-squares rigid alignment of those pairs via Kabsch, and applies that transform to A, thus A’s skull geometry is brought into B’s coordinate.

\paragraph{Lens Transfer}

\begin{figure}[ht]
    \centering
    \includegraphics[width=0.5\textwidth]{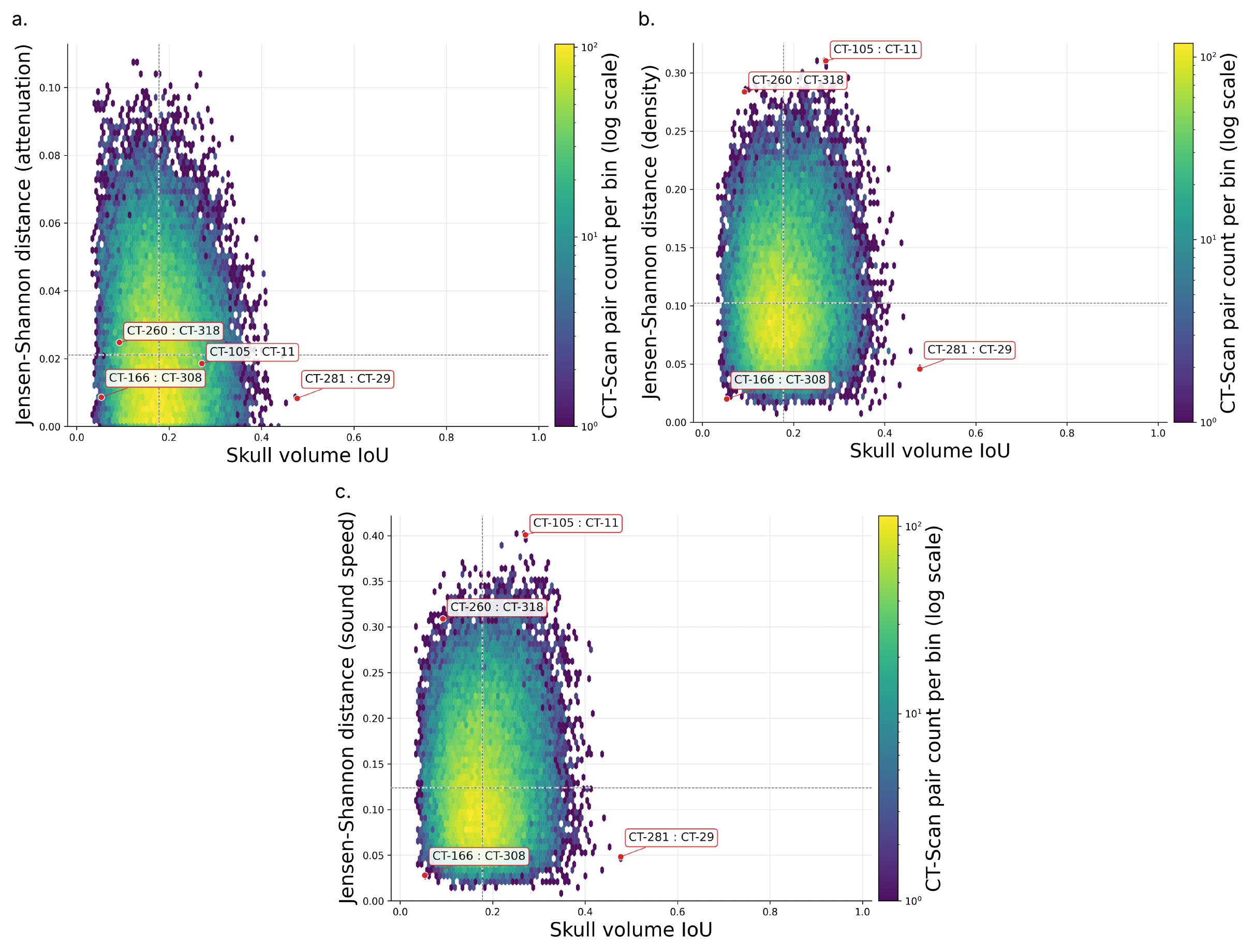}
    \caption{Density plots of Jensen-Shannon(y-axis) by IoU(x-axis) is plotted for all pairs with a. attenuation, b. density, and c. speed}
    \label{fig:physical_comp}
\end{figure}

\begin{table}[ht]
\centering
\resizebox{\columnwidth}{!}{%
\begin{tabular}{cccc}
\hline
\textbf{CT-A} & \textbf{CT-B} & \textbf{Misfocus Dist. (center)} & \textbf{Misfocus Dist. (off-center)} \\ \hline
105 & 11 & 11.172 & 22.706 \\
260 & 318 & 21.995 & 29.593 \\
166 & 308 & 12.189 & 19.640 \\
281 & 29 & 7.673 & 21.269 \\ \hline
\end{tabular}%
}
\caption{Misfocus distances(mm) for central and off-center focusing across different CT pairs.}
\label{tab:lens_transfer_misfocus}
\end{table}

By computing the IoU and Jensen-Shannon distance between different CT pairs, we are able to model the relationship between acoustic and structural properties by plotting skull volume IoU on the x-axis and Jensen-Shannon distance on y-axis. Figure~\ref{fig:physical_comp} is partitioned into four quadrants each holding CT-pairs that represent the same relationship between skull and acoustic properties. We choose one CT-pair(A,B) from each quadrant((CT-105,CT-11), (CT-260, CT-318), (CT-166,CT-308), (CT-281, CT-29) for a lens-transfer comparison. For lens transfer, an acoustic lens is fit and optimized for CT-A, then placed on CT-B and misfocused is calculated. A point on CT-A is correlated to a point on CT B by performing Iterative Closest Point(ICP) rigid alignment as described in the Appendix. Table~\ref{tab:lens_transfer_misfocus} shows that a lens cannot be accurately transferred even if a pair of CT-scans are acoustically and structurally similar or dissimilar. 

Table ~\ref{tab:lens_transfer_misfocus} shows that the acoustically and structurally similar pair(CT-105 and CT-11) have the least misfocus distance for center focusing while the acoustically and structurally disimilar pair(CT-260 and CT-318) had the largest misfocus for center and off-center focus. In the brain, the misfocusing of \textbf{29.593mm} is the difference between the \textbf{thalamus} and \textbf{cerebral cortex}. Thus, acoustic lens personalization, especially for millimeter-scale focusing is critical and necessary for heads that are maybe acoustically and structurally similar.
\subsection{CT-to-Acoustic Conversion}
\label{app:ct}

Hounsfield units (HU) are extracted from the DICOM data using
SimpleITK~\cite{lowekamp2013design}. HU measures the degree to which a
material attenuates X-rays, and density is recovered from HU using k-Wave's
\texttt{hounsfield2density}~\cite{treeby2010k}, which implements the
calibration of~\cite{schneider1996calibration}. Sound speed follows the
linear density mapping of~\cite{mast2000empirical},
\begin{equation}
    c = 1.33\rho + 167,
    \label{eq:mast}
\end{equation}
with $\rho$ in $\mathrm{kg/m^3}$ and $c$ in $\mathrm{m/s}$.

No agreed-upon HU-to-attenuation conversion exists, so we assign constant
attenuation per tissue class: $13.3\,\mathrm{dB/cm/MHz}$ for
skull~\cite{pinton2012attenuation}, $0.55\,\mathrm{dB/cm/MHz}$ for brain
soft tissue~\cite{hoffmeister2025fundamental}, and
$0.02\,\mathrm{dB/cm/MHz}$ for cerebrospinal fluid, which is approximately
99\% water and exhibits negligible ultrasonic absorption relative to brain
tissue~\cite{slominski2023influence}. Per-slice preprocessing removes
imaging artifacts and extracts the head region.

\subsection{Domain Configuration}
\label{app:domain}

The preprocessed scan is placed into the simulator and all axes are then
cropped to fit the head, so that the optimization accounts for whole-head
physics rather than a truncated region. A perfectly matched layer of 10 grid
points absorbs outgoing waves at the domain boundaries and prevents
reflections back into the simulation. Every domain is resampled to
$0.4\,$mm, which keeps resolution consistent across patients and makes lens
transfer between patients well-defined. For all experiments the head is
rotated $90^{\circ}$ counterclockwise so the source propagates along the
$x$-axis. A $1\,$mm water bed between lens and head models ultrasound gel,
and the surrounding medium is an absorbing layer emulating free space.

Transducers are radial with a Gaussian aperture of radius $r$, and are
embedded directly into the lens to eliminate intermediate media between
source and lens. We consider two configurations: a single transducer
centered on the lens, and an $n \times n$ grid whose element radius is
maximized subject to non-overlap at the prescribed head coverage. On a
$1\times1$ grid at $60\%$ coverage the single element has radius $25\,$mm;
element radius shrinks as $n$ grows.

\paragraph{Main Results Configuration:}
For the main results, we use choose three acoustically varying CT acoustic models from our Analysis of Inter-Patient Variability: CT-105, CT-166, and CT-260. A 10x10 transducer grid with 500 kHz each. The grid of transducers are embedded in the first layer of the lens. The lens is centered on the head and is set to cover 60\% of the surface. The focus is a sphere with a radius of 4 mm and is placed at the center of the head for one experiment and placed off-center for another experiment.

\subsection{Baseline Formulations}
\label{app:baselines}

\subsection{Homogeneous Phased-Array Transducer}
The homogeneous PAT computes phase delays assuming a uniform water medium.
For element $i$ at $\mathbf{r}_i$ and focus $\mathbf{r}_f$, the Euclidean
distance is $d_i = \lVert \mathbf{r}_f - \mathbf{r}_i \rVert$. Taking the
nearest element as reference, $d_{\mathrm{ref}} = \min_j d_j$, the
excitation is
\begin{equation}
    \phi_i = -k\,(d_i - d_{\mathrm{ref}}),
    \qquad
    a_i = e^{j\phi_i},
    \label{eq:homog_pat}
\end{equation}
with $k = 2\pi f / c_{\mathrm{water}}$ and
$c_{\mathrm{water}} = 1480\,\mathrm{m/s}$. Amplitudes are unit
($|a_i| = 1$) for consistency across methods.

\paragraph{Heterogeneous Green's Function Phase Computation}
The heterogeneous PAT~\cite{aubry2003experimental,marquet2009non} uses the
acoustic model derived from the patient CT. For each element $i$, the
frequency-domain Helmholtz equation is solved with only that element active,
giving pressure field $p_i(\mathbf{x})$. The complex response at the target
region $\Omega_f$ is the average over the focus volume,
\begin{equation}
    G_i = \frac{1}{|\Omega_f|} \sum_{\mathbf{x} \in \Omega_f} p_i(\mathbf{x})
        = |G_i| e^{j\phi_i},
    \label{eq:greens}
\end{equation}
where $|G_i|$ is the response magnitude and $\phi_i$ the phase accumulated
by propagation through the heterogeneous medium. Aberration is compensated
by phase conjugation,
\begin{equation}
    a_i = \frac{\overline{G_i}}{|G_i|} = e^{-j\phi_i},
    \label{eq:conjugation}
\end{equation}
with $j = \sqrt{-1}$. Only phase is corrected; amplitude is fixed to unity.

\paragraph{Thickness-Based Lens}
Thickness-based lenses~\cite{jimenez2019holograms,hu2022binary} convert a
required phase into an equivalent thickness profile. A slab of thickness $t$
introduces phase delay
\begin{equation}
    \phi = -\omega t \left( \frac{1}{c_{\mathrm{lens}}} - \frac{1}{c_{\mathrm{water}}} \right),
    \label{eq:slab}
\end{equation}
so the required thickness is
\begin{equation}
    t = -\frac{\phi}{\omega\left(\frac{1}{c_{\mathrm{lens}}} - \frac{1}{c_{\mathrm{water}}}\right)}.
    \label{eq:thickness}
\end{equation}
Since phase is periodic, thickness is wrapped to a single $2\pi$ cycle,
\begin{equation}
    t_{2\pi} = \frac{2\pi}{\omega \left| \frac{1}{c_{\mathrm{lens}}} - \frac{1}{c_{\mathrm{water}}} \right|},
    \qquad
    t \leftarrow t \bmod t_{2\pi},
    \label{eq:wrap}
\end{equation}
with $t \in [0, t_{\max}]$ for maximum allowable thickness $t_{\max}$. The
continuous thickness map is discretized onto the simulation grid: for each
$(y,z)$ coordinate the voxel count along the propagation direction is
\begin{equation}
    n = \max\!\left(0,\, \left\lceil \frac{t}{\Delta x} \right\rceil \right).
    \label{eq:thickness_voxels}
\end{equation}

\paragraph{Sigmoid Lens}
Density-based topology optimization
methods~\cite{stanziola2023physics,lee2024machine} represent material properties
directly at each voxel,
\begin{align}
    \rho(\mathbf{x}) &= \rho_{\text{water}} + \alpha(\mathbf{x})(\rho_{\max} - \rho_{\text{water}}), \\
    c(\mathbf{x})    &= c_{\text{water}}    + \alpha(\mathbf{x})(c_{\max}    - c_{\text{water}}),
\end{align}
with $\alpha(\mathbf{x}) = \sigma(\mathbf{x})$ for the logistic sigmoid
$\sigma(x) = 1/(1 + e^{-x})$, water properties
$\rho_{\text{water}} = 1000\,\mathrm{kg/m^3}$ and
$c_{\text{water}} = 1480\,\mathrm{m/s}$, and user-defined maxima
$\rho_{\max}$, $c_{\max}$. This baseline shares our continuous
parametrization but imposes no fabrication constraint. Then to make it binary, a hard threshold is performed, exactly what was performed in ~\cite{stanziola2023physics}:
\begin{equation}
    \bar{\alpha}(\mathbf{x}) = \mathcal{H}(\alpha(\mathbf{x}) - \tau) = 
    \begin{cases} 
    1, & \text{if } \alpha(\mathbf{x}) \ge \tau, \\
    0, & \text{otherwise},
    \end{cases}
\end{equation}
where $\mathcal{H}(\cdot)$ denotes the Heaviside step function and $\tau \in (0, 1)$ is a predefined binarization threshold (set to $\tau = 0.5$).

\subsection{Supplementary Visualizations}

\begin{figure}[ht]
    \centering
    \includegraphics[width=0.5\textwidth]{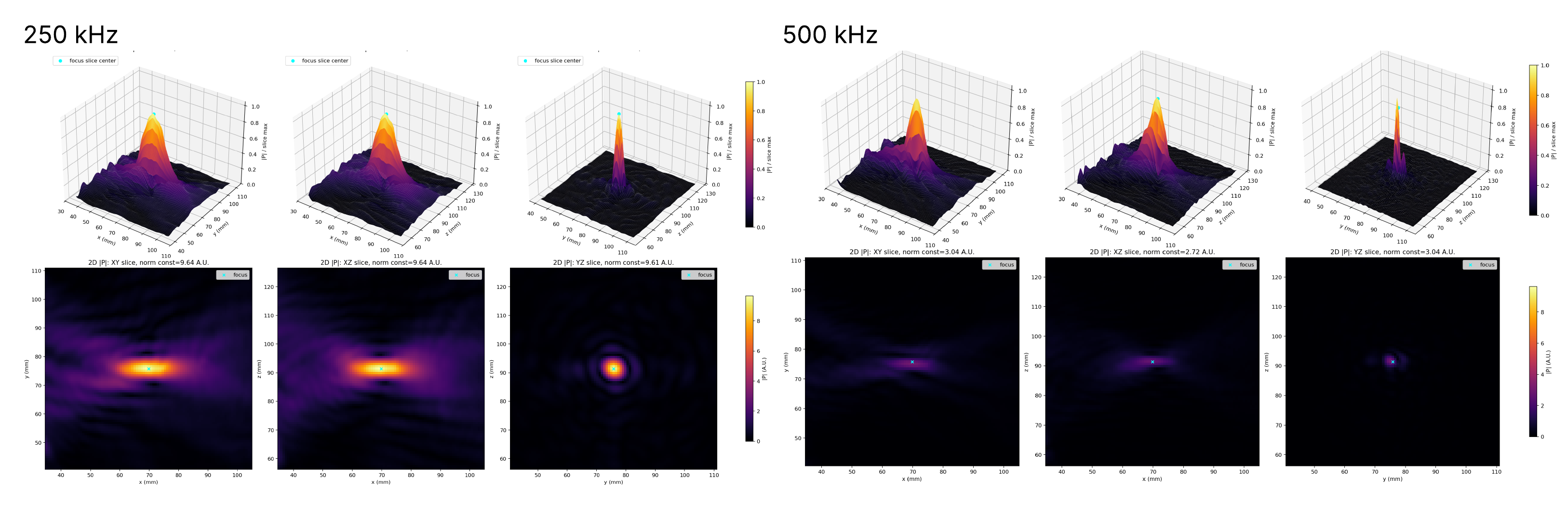}
    \caption{Normalized pressure landscape (XY, YZ, and XZ slices) for 250~kHz (left) and 500~kHz (right).}
    \label{fig:250v500}
\end{figure}

\begin{figure}[ht]
    \centering
    \includegraphics[width=0.5\textwidth]{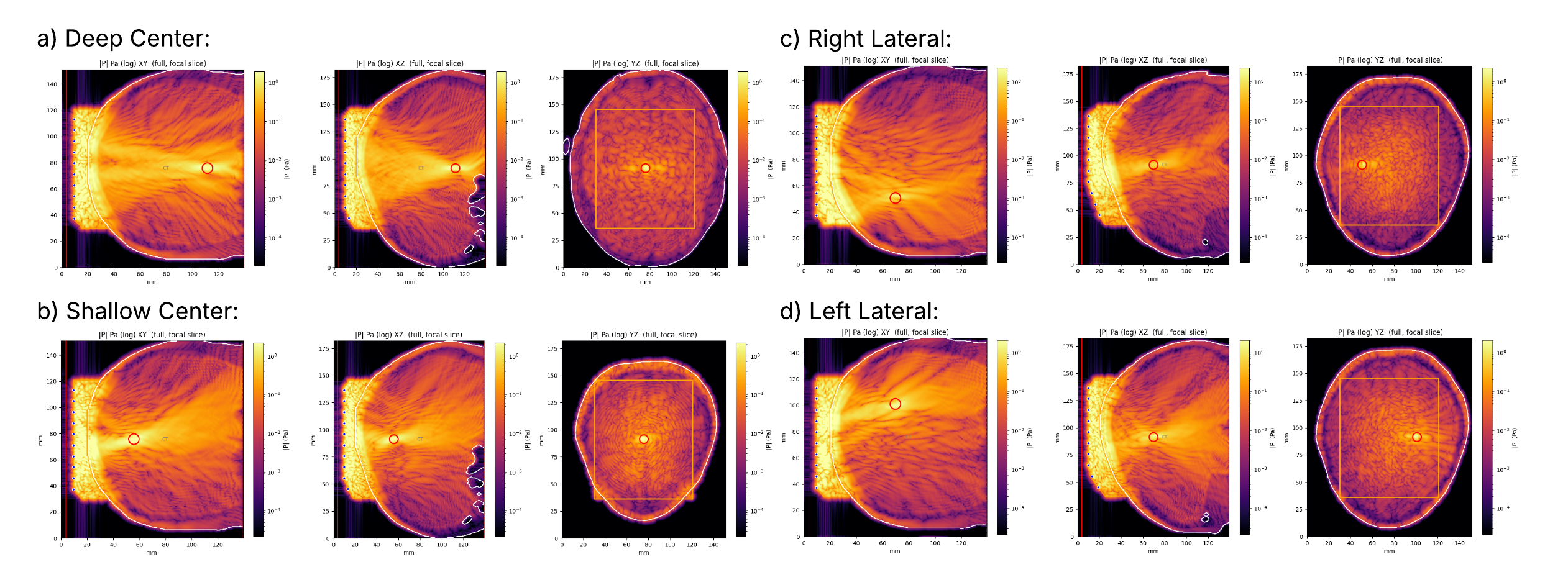}
    \caption{Four different focus locations: a) Deep Center (top left), b) Shallow Center (bottom left), c) Right Lateral (top right), and d) Left Lateral (bottom right).}
    \label{fig:fourdiflocs}
\end{figure}

\subsection{SIREN parameter ablation}

In this section, we show our ablation performed in order to find optimal parameters. Main results were performed on CT-105, however, in this ablation study, CT-120 was used. 

\begin{table}[h!]
\centering
\small
\begin{tabularx}{\columnwidth}{@{} >{\footnotesize\raggedright\arraybackslash}X c c c c c@{}}
\hline
\textbf{$\omega$} & \textbf{Overlap (\%)} & \textbf{Ratio} & \textbf{Shift (mm)} & \textbf{Peak P.} & \textbf{RMS P.} \\ \hline
30 & 22.9 & 1.835 & 0.89 & 2.588 & 0.750 \\
60 & 23.8 & 1.830 & 0.69 & 3.315 & 0.968 \\
90 & 28.0 & 1.815 & 0.69 & 3.218 & 1.008 \\
120 & 28.0 & 1.793 & 0.89 & 2.777 & 0.877 \\ \hline
\end{tabularx}
\caption{Ablation results for SIREN INR $\omega$}
\label{tab:results_sirenomega}
\end{table}

We first experiment with the $\omega$ in the SIREN INR, which controls the frequency of the sinusoidal activations and also the INR's ability to represent high-frequency spatial variations. As illustrated in Table~\ref{tab:results_sirenomega}, increasing $\omega$ from 30 to 90 improves the overlap of 22.9\% to 28.0\% while maintaining the focal shift. The peak and RMS pressure also increase until $\omega=90$. $\omega=120$ does not improve overlap; instead, it reduces both peak and RMS pressure while slightly increasing the focal shift. Overall, we choose $\omega=90$, which provides the best trade-off between focal accuracy and pressure concentration.

\begin{table}[h!]
\centering
\small
\begin{tabularx}{\columnwidth}{@{} >{\footnotesize\raggedright\arraybackslash}X c c c c c@{}}
\hline
\textbf{$\tau$} & \textbf{Overlap (\%)} & \textbf{Ratio} & \textbf{Shift (mm)} & \textbf{Peak P.} & \textbf{RMS P.} \\ \hline
0.005 & 37.9 & 1.763 & 0.98 & 1.264 & 0.450 \\
0.01 & 31.0 & 1.747 & 2.08 & 1.484 & 0.479 \\
0.05 & 28.7 & 1.817 & 0.69 & 2.305 & 0.718 \\
0.1 & 26.5 & 1.795 & 1.13 & 2.270 & 0.699 \\ \hline
\end{tabularx}
\caption{Ablation results for Gumbel temperature $\tau$}
\label{tab:results_gumbeltemp}
\end{table}

The Gumbel temperature $\tau$ controls the sharpness of the continuous approximation of the binary occupancy field. As shown in Table~\ref{tab:results_gumbeltemp}, the lowest temperature achieves the highest overlap, but at the expense of lower peak and RMS pressure. Increasing $\tau$ generally reduces the overlap while improving the pressure metrics, with $\tau=0.05$ achieving the highest ratio (1.817) and substantially higher peak and RMS pressure. Since accurate and high-concentration focusing is the primary objective, we select $\tau=0.05$ for all subsequent experiments.

\begin{table}[h!]
\centering
\small
\begin{tabularx}{\columnwidth}{@{} >{\footnotesize\raggedright\arraybackslash}X c c c c c@{}}
\hline
\textbf{$\tau$} & \textbf{Overlap (\%)} & \textbf{Ratio} & \textbf{Shift (mm)} & \textbf{Peak P.} & \textbf{RMS P.} \\ \hline
0.01 & 35.9 & 1.756 & 1.26 & 1.441 & 0.496 \\
0.05 & 29.9 & 1.771 & 1.33 & 1.560 & 0.523 \\
0.1 & 32.5 & 1.781 & 0.89 & 1.579 & 0.523 \\
0.25 & 30.4 & 1.765 & 0.98 & 1.524 & 0.496 \\ \hline
\end{tabularx}
\caption{Ablation results for Sigmoid Gate hyper-parameter $\tau_g$}
\label{tab:results_gatetau}
\end{table}

The sigmoid gate temperature, $\tau$, controls the sharpness of the gating function used to map the continuous occupancy values toward binary states. Smaller values produce a steeper transition, while larger values yield a smoother gate. Illustrated in Table~\ref{tab:results_gatetau}, increasing $\tau_g$ generally leads to a slight reduction in overlap, although the ratio and pressure metrics improve marginally. Due to its superiority in all five metrics, we choose $\tau_g=0.01$ in all subsequent experiments.

\begin{table}[h!]
\centering
\small
\begin{tabularx}{\columnwidth}{@{} >{\footnotesize\raggedright\arraybackslash}X c c c c c@{}}
\hline
\textbf{$\lambda$} & \textbf{Overlap (\%)} & \textbf{Ratio} & \textbf{Shift (mm)} & \textbf{Peak P.} & \textbf{RMS P.} \\ \hline
0.01 & 32.0 & 1.736 & 1.70 & 1.429 & 0.4703 \\
0.02 & 29.6 & 1.751 & 1.69 & 1.576 & 0.498 \\
0.05 & 38.1 & 1.760 & 2.40 & 1.366 & 0.487 \\
0.075 & 32.1 & 1.688 & 2.20 & 1.498 & 0.454 \\ \hline
\end{tabularx}
\caption{Ablation results for Printability loss term hyper-parameter $\lambda$}
\label{tab:results_lambdaprint}
\end{table}

The final tunable hyperparameter is the printability loss weight $\lambda$, which balances the optimization objective between acoustic focusing and printability. A larger $\lambda$ will place greater emphasis on satisfying printability, while a smaller value prioritizes the acoustic objective. As shown in Table~\ref{tab:results_lambdaprint}, $\lambda=0.01$ achieves the peak pressure and the lowest shift, indicating the best overall focal quality. Thus, we set $\lambda=0.02$ for all subsequent experiments.

\subsection{Supplementary Results}

\begin{table}[h!]
\centering
\small
\resizebox{\columnwidth}{!}{% 
\begin{tabular}{@{} l *{5}{c} @{}} 
\hline
\textbf{Method} & 
\begin{tabular}{@{}c@{}}\textbf{Overlap} \\ \textbf{(\%)}\end{tabular} & 
\begin{tabular}{@{}c@{}}\textbf{Ratio} \\ \textbf{(R/2R)}\end{tabular} & 
\begin{tabular}{@{}c@{}}\textbf{Shift} \\ \textbf{(mm)}\end{tabular} & 
\begin{tabular}{@{}c@{}}\textbf{Peak P.} \\ \textbf{(A.U)}\end{tabular} & 
\begin{tabular}{@{}c@{}}\textbf{RMS P.} \\ \textbf{(A.U)}\end{tabular} \\ \hline
Unconstrained Ref. & 41.5 & \textbf{1.822} & 1.55 & \textbf{1.29} & \textbf{0.469} \\
\sysName & \textbf{47.3} & 1.796 & 0.94 & 0.931 & 0.347 \\
Homo. PAT & 41.8 & 1.327 & 7.87 & 0.175 & 0.0337 \\
Hetero. PAT & 23.8 & 1.688 & \textbf{0.53} & 0.414 & 0.117 \\
Thickness & 31.5 & 0.986 & 5.52 & 0.100 & 0.0309 \\
Sigmoid & 42.8 & 1.586 & 2.58 & 0.331 & 0.119 \\ \hline
\end{tabular}%
}
\caption{Main Results for Off-Center Focusing. Average of CT-105, CT-166, and CT-260 results.}
\label{tab:results_offcenter}
\end{table}

Additionally, we include results for off-centering, comparising against all other baselines. Table~\ref{tab:results_offcenter} illustrates the same results as the center results in  the paper, however, the Heterogeneous PAT has a better shift than unconstrained and AcoustiCraft. Heterogeneous PAT achieves the smallest focal shift because it compensates for spatial variations in skull acoustic properties, reducing phase aberrations that would otherwise displace the focus. However, while the focal location is more accurate, the resulting energy is distributed over a broader region, leading to substantially lower overlap and reduced pressure metrics.

\paragraph{Effect of different number of transducers}
\begin{table}[h!]
\centering
\small % Slightly smaller font to ensure comfortable fit
\begin{tabularx}{\columnwidth}{@{}X X c c c c@{}}
\hline
\textbf{Grid size} & \textbf{Overlap (\%)} & \textbf{Ratio} & \textbf{Shift (mm)} & \textbf{Peak P.} & \textbf{RMS P.} \\ \hline
1 & 55.9 & 1.760 & 4.04 & 1.771 & 0.746 \\
2 & 40.1 & 1.704 & 1.44 & 0.503 & 0.169 \\
4 & 28.6 & 1.814 & 0.57 & 1.997 & 0.623 \\
6 & 26.8 & 1.819 & 0.57 & 3.508 & 1.065 \\
8 & 27.1 & 1.814 & 0.69 & 3.205 & 0.978 \\
10 & 32.5 & 1.848 & 0.98 & 2.34 & 0.763 \\ \hline
\end{tabularx}
\caption{Results for different size grids of transducers.}
\label{tab:results_numtransducers}
\end{table}

Transducers are laid on a grid of size nxn where a transducer on a 1x1 grid will have 25mm in radius and as the number of transducers increases, the size of each transducer decreases in size. This is because the lens is set to cover a constant 60\% of the patients head. The radius of each transducer is determined as the maximum size that will result in a evenly spaced grid while constraint to a 60\% head coverage.

Table~\ref{tab:results_numtransducers} displays that as the number of transducers increase, the volume overlap decreases since more pressure is being distributed among the focus region. Additionally, focus shift decreases as more transducers help target the peak towards intended focus. Peak and RMS pressure both plateau since the power at the focus will stay near constant due to the decrease in transducer size.

\paragraph{Effect of different lens sizes}

\begin{table}[h!]
\centering
\small
\begin{tabularx}{\columnwidth}{@{}X c c c c c@{}}
\hline
\textbf{\%} & \textbf{Overlap (\%)} & \textbf{Ratio} & \textbf{Shift (mm)} & \textbf{Peak P.} & \textbf{RMS P.} \\ \hline
20 & 89.7 & 1.546 & 4.44 & 0.6089 & 0.3014 \\
40 & 45.7 & 1.764 & 0.57 & 1.998 & 0.7804 \\
60 & 32.5 & 1.848 & 0.98 & 2.34 & 0.763  \\
80 & 20.4 & 1.858 & 0.89 & 3.573 & 0.9510 \\
100 & 19.2 & 1.856 & 0.89 & 3.760 & 0.9837 \\ \hline
\end{tabularx}
\caption{Results for varying transducer head coverage percentages.}
\label{tab:results_headcoverage}
\end{table}

Lens size is defined as the percentage of the head that is covered by the lens. Acoustic focal accuracy and concentration increases as lens size increases due to more area coverage by transducers. Larger lens sizes gives more freedom to the INR to create a better-performing lens.

\subsection{Results Replication:}
For Results replication, please look at the README.md provided with the code materials.

\subsection{Hyperparameters:}

\begin{table}[h!]
\centering
\small
\begin{tabularx}{\columnwidth}{@{}X c c@{}}
\hline
\textbf{Hyper-parameter} & \textbf{Unconstrained} & \textbf{Printable} \\ \hline
Frequency & 500 kHz & 500 kHz \\
Optimizer & Adam & Adam \\
Optimizer steps & 30 & 30 \\
Learning rate & $5\times10^{-4}$ & $5\times10^{-4}$ \\
SIREN hidden units & 128 & 128 \\
SIREN depth & 4 & 4 \\
SIREN $\omega_0$ & 100 & 100 \\
Lens sound speed & 2500 m/s & 2500 m/s \\
Lens density & 1200 kg/m$^3$ & 1200 kg/m$^3$ \\
GMRES iterations & 400 & 400 \\
GMRES restart & 10 & 10 \\
Lens--skull gap & 1 mm & 1 mm \\
Transducer grid & $10 \times 10$ & $10 \times 10$ \\
Gumbel sigmoid & -- & Enabled \\
Gumbel temperature & -- & 0.01 \\
Straight-through estimator & -- & Enabled \\
Gate temperature & -- & 0.1 \\
Slicer penalty ($\lambda$) & -- & 0.02 \\
Build axis & -- & $z$ \\
Maximum overhang angle & -- & $45^\circ$ \\ \\\hline
\end{tabularx}
\caption{Optimization hyperparameters used for unconstrained and printable lens optimization. Parameters not applicable to the unconstrained method are denoted by ``--''.}
\label{tab:hyperparameters}
\end{table}

\subsection{Hyperparameter tags:}

\begin{table}[h!]
\centering
\small
\begin{tabularx}{\columnwidth}{@{}X X@{}}
\hline
\textbf{Result} & \textbf{Modification} \\ \hline
Table~2 (Frequency study) & Change \texttt{--frequency}. \\
Table~3 (Focus location study) & Change \texttt{--focus-location}. \\
Table~7 (SIREN initialization) & Change \texttt{--siren-omega0}. \\
Table~8 (Gumbel temperature) & Change \texttt{--gumbel-temperature}. \\
Table~9 (Seed gate temperature) & Change \texttt{--seed-gate-temperature}. \\
Table~10 (Slicer penalty) & Change \texttt{--slicer-penalty}. \\
Table~11 (Lens transfer) & Run \texttt{lens\_transfer.py}. \\
Table~12 (Transducer size study) & Change \texttt{--transducer-grid-n}. \\
\hline
\end{tabularx}
\caption{Instructions for reproducing the quantitative results reported in the paper. Unless otherwise specified, all experiments use the default unconstrained configuration. Printable results are obtained by enabling the printability arguments listed in Table~\ref{tab:hyperparameters}.}
\label{tab:reproduction}
\end{table}

\subsection{Dataset Description:}
We choose to use Qure.aiCQ500 Head CT-scan dataset~\cite{chilamkurthy2018development} . It is available both on the Qure.ai website and Kaggle. We use 290 Qure.ai CQ500 Head CT-scan’s from Qure.ai CQ500 dataset for our acoustic study in the Appendix and 3 for our main results(CT-105, CT-166, CT-260). All other experiments use CT-105.

\end{document}